\pdfoutput=1
\documentclass[letterpaper, 10 pt, conference]{ieeeconf}  

\IEEEoverridecommandlockouts                              

\newcommand{\qq}[1]{VISION}

\usepackage{graphicx} 
\usepackage{multirow}
\usepackage{multicol}
\usepackage{array,booktabs}
\usepackage[skins,theorems]{tcolorbox}
\usepackage{bm}
\usepackage{url}
\usepackage{tabularx}
\usepackage{xcolor}
\usepackage{caption}
\usepackage[font=small,labelfont=bf]{caption}
\usepackage{subcaption}
\usepackage[font=small,labelfont=bf]{subcaption}
\usepackage{float}
\usepackage{soul}
\usepackage{graphicx}
\usepackage{etoolbox}
\usepackage{hyperref}

\usepackage{xcolor}
\usepackage{minted}

\definecolor{codebg}{rgb}{0.95,0.95,0.92}

\setminted{
  breaklines,
  autogobble,
  bgcolor=codebg,
  frame=single,        
  framesep=3mm,        
  fontsize=\scriptsize
}
\usepackage[normalem]{ulem}
\usepackage{hyperref}
\usepackage{cleveref}
\crefname{figure}{Fig.}{Figs.}
\Crefname{figure}{Figure}{Figures}

\crefname{table}{Tab.}{Tabs.}
\Crefname{table}{Table}{Tables}

\crefname{equation}{Eq.}{Eqs.}
\Crefname{equation}{Equation}{Equations}

\title{\LARGE \bf
Language-in-the-Loop Culvert Inspection on the Erie Canal

}

\newcommand{\insertfig}{
\includegraphics[trim=1.5cm 1.25cm 1.25cm 1.25cm, clip=true,width=\textwidth]{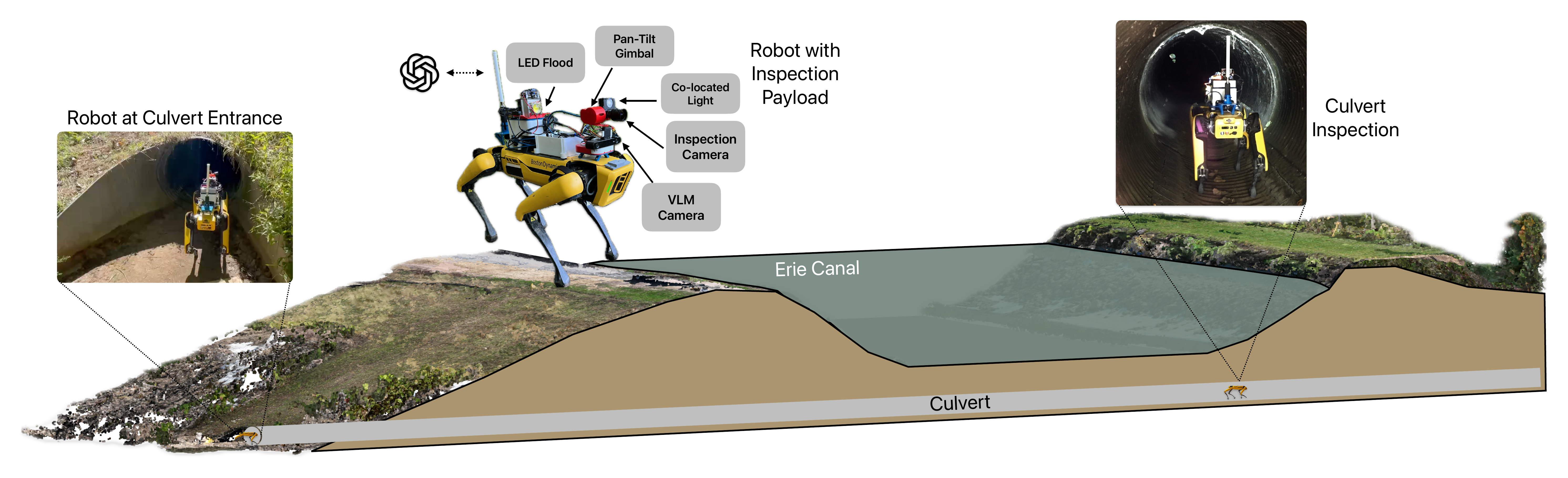}
\captionof{figure}{
Schematic of the Erie Canal culvert inspection setup. The cross-section of the canal and culvert is drawn to scale, derived from a 3D reconstruction of the site, illustrating the buried drainage conduit, a.k.a. the culvert, beneath the canal embankment. Culverts provide critical drainage but their confined geometry and location make inspection difficult. A Boston Dynamics Spot quadruped, outfitted with a custom inspection payload (pan–tilt gimbal, inspection camera with co-located light, VLM proposal camera, and auxiliary LED flood), is deployed at the culvert entrance and traverses the interior. Insets show the robot at the portal and during inspection runs inside the conduit. Only the robot payload illustration is not to scale; all other canal and culvert geometry is based on the site reconstruction. The system autonomously navigates through the 66 m long, 1.2 m diameter culvert, capturing targeted images for structural condition assessment.
}
\label{fig:intro-splash}
\vspace{-0.5cm}
}

\makeatletter
\newif\ifdid@splash \did@splashfalse
\apptocmd{\@maketitle}{%
  \ifdid@splash\else
    \centering\insertfig
    \global\did@splashtrue
  \fi
}{}{}
\makeatother

\author{Yash Turkar$^{*}$, Yashom Dighe$^{*}$, Karthik Dantu 
\\ Department of Computer Science and Engineering, University at Buffalo, Buffalo, NY
\thanks{$^{*}$Authors contributed equally. \href{https://droneslab.github.io/vision/}{{\color{blue}Project page}}. 
This work is in collaboration with the New York Canal Corporation (NYCC)}}

\begin{document}
\maketitle

\thispagestyle{empty}
\pagestyle{empty}

\begin{abstract}

Culverts on canals such as Erie Canal built originally in 1825 require frequent inspections to ensure safe operation. 
Human inspection of culverts is challenging due to age, geometry, poor illumination, weather and lack of easy access. 
We introduce \textbf{\qq{}}, an end-to-end, language-in-the-loop autonomy system that couples a web-scale vision–language model (VLM) with constrained viewpoint planning for autonomous inspection of culverts. Brief prompts to the VLM solicit open-vocabulary ROI proposals with rationales and confidences, stereo depth is fused to recover scale, and a planner—aware of culvert constraints commands repositioning moves to capture targeted close-ups. Deployed on a quadruped in Culvert under the Erie canal, \qq{} closes the see→decide→move→re-image loop on-board and produces high-resolution images for detailed reporting without domain-specific fine-tuning. In an external evaluation by New York Canal Corporation personnel, initial ROI proposals achieved 61.4\% agreement with subject-matter experts, and final post-re-imaging assessments reached 80\%, indicating that \qq{} converts tentative hypotheses into grounded, expert-aligned findings.

\end{abstract}

\section{Introduction}
\label{sec:intro}


Canals are artificial waterways built for drainage management and water transport. Largely built in the 19th and 20th century, there are over 4000 miles of canals in the US \cite{britannicaCanalsInland}. Erie canal is one such canal that is 363 miles in length and runs east-west between the Hudson river and Lake Erie. It was initially built in 1825 with several expansions including a major one between 1905 and 1918. Culverts are small cylindrical passages under the canal typically built for draining water after wet spells. The erie canal has over 350 culverts. Most of them were built using concrete, metal or stone. However, over time, these culverts are subject to structural deterioration, as reflected in New York Canal Corporation (NYCC) inspection records. Typical defects include surface corrosion, spalling, and seepage, any of which can escalate to culvert failure with serious consequences for adjacent communities and ecosystems. Currently, these culverts are inspected manually, requiring personnel to enter confined underground spaces in remote locations. Such inspections pose significant risks: the culverts lie beneath the canal in restricted, poorly ventilated environments, where visibility and air quality are uncertain,  and structural stability cannot be guaranteed. Reaching these sites is often difficult due to steep slopes, uneven terrain, and waterlogged conditions, all of which increase the likelihood of accidents. These hazards make manual inspection both dangerous and impractical, motivating the development of automated alternatives. Legged robots offer a promising alternative~\cite{Singh2025} as they can descend embankments, wade through shallow water, and carry multi-modal sensors for close-range imaging.

\begin{figure}
    \centering
    \includegraphics[width=\linewidth]{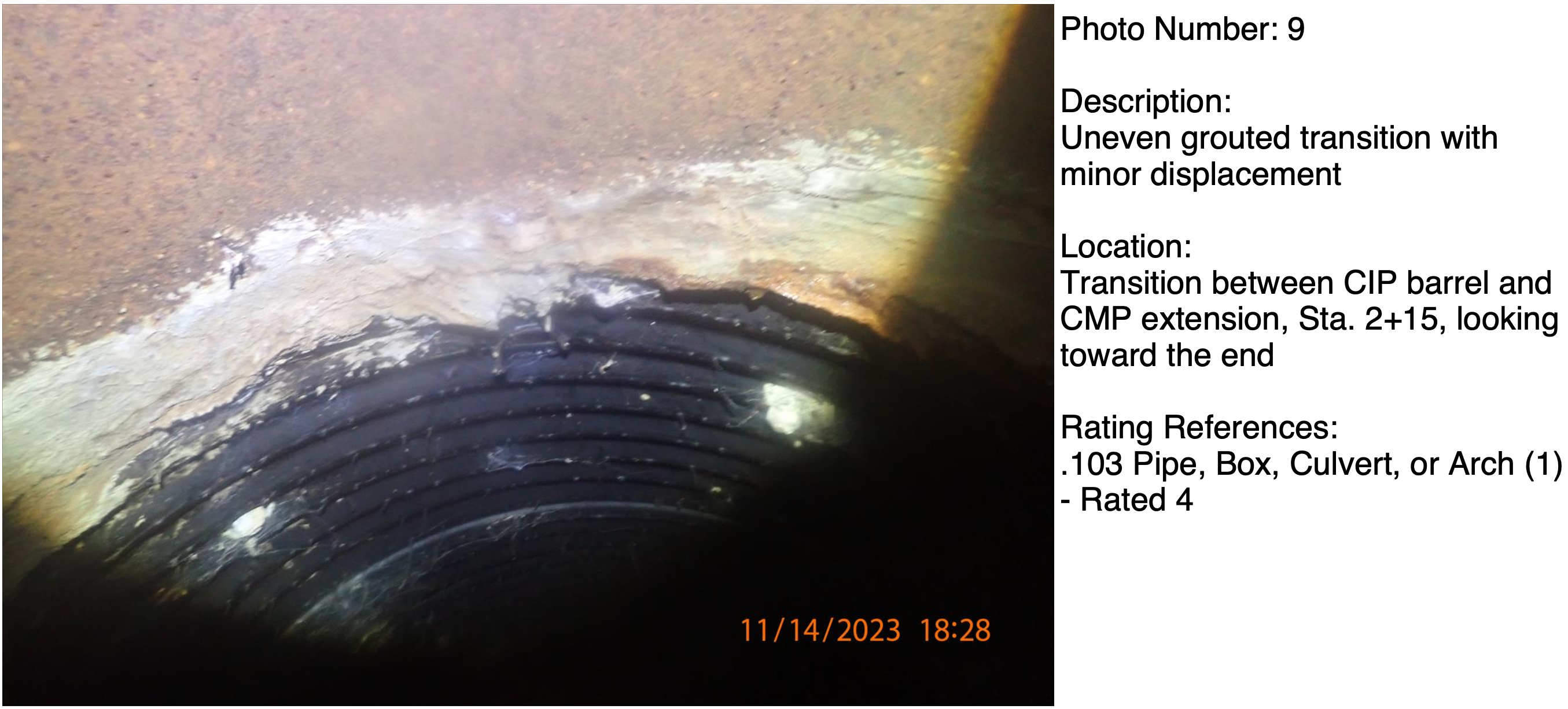}
    \caption{
    Sample from an official Inspection report by NYCC}
    \label{fig:nycc-report}
    \vspace{-0.7cm}
\end{figure}

\begin{figure*}
    \centering
    \includegraphics[width=\linewidth]{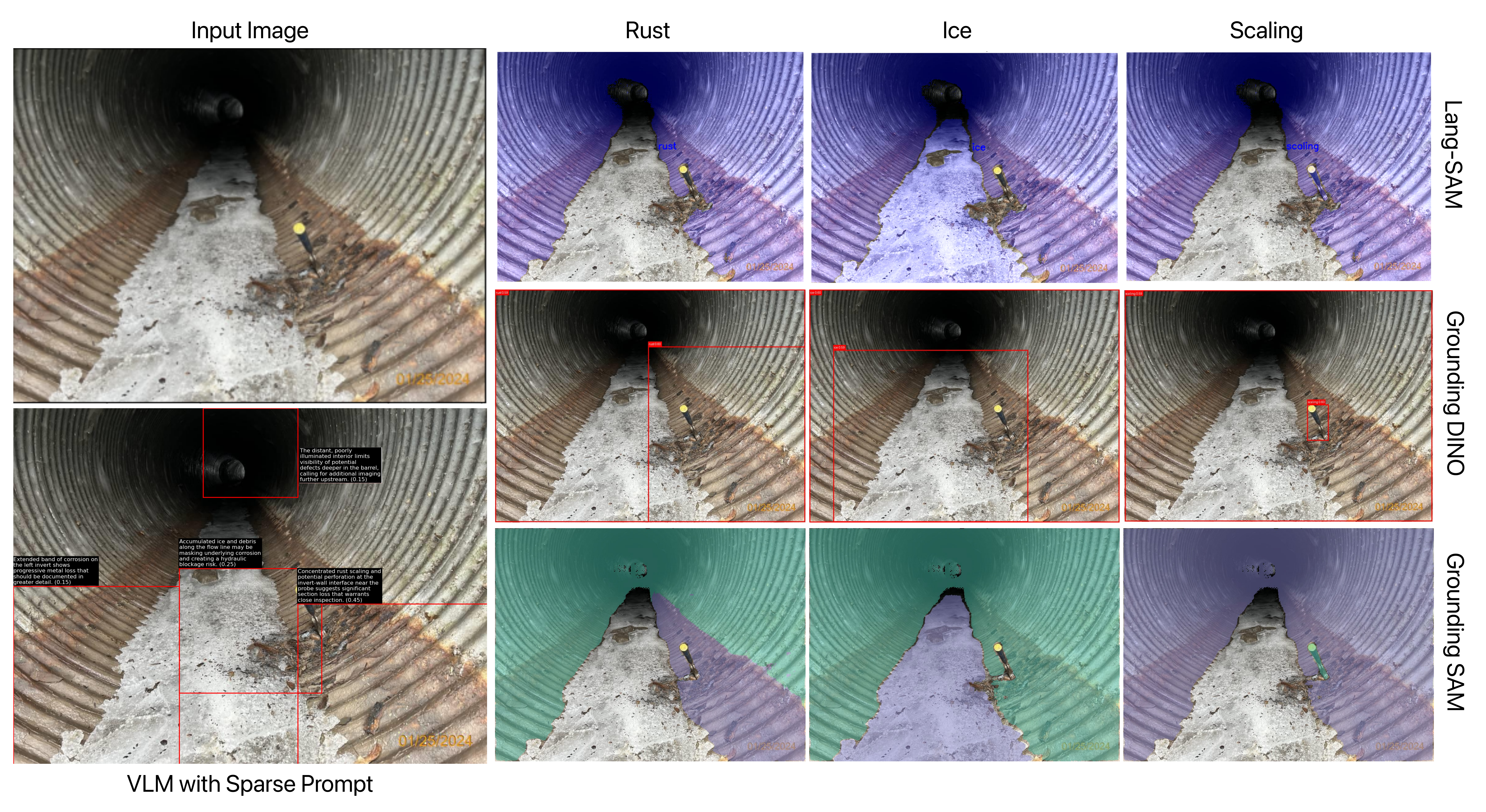}
    \caption{Qualitative comparison across degradations and baselines. Left: input culvert image and our VLM with a sparse prompt producing ROI proposals (red boxes with reasons) and calibrated follow-up probabilities (shown in parentheses; sum to 1.00). Right: results for three degradation modes—rust, ice/debris, and scaling—under three open-vocabulary baselines: Lang-SAM, Grounding DINO, and Grounding SAM. Lang-SAM often over-segments large portions of the barrel; Grounding DINO yields coarse boxes; Grounding SAM converts those boxes to masks but still spreads labels broadly. Our method localizes discrete issues (e.g., rust staining at corrugations, ice/debris on the invert, mineral scaling near the crown) to guide targeted re-imaging.}
    \vspace{-0.7cm}
    \label{fig:qualitative}
\end{figure*}

However, despite the ability of legged robots to traverse such constrained environments, culvert inspection remains a challenging task due to unreliable defect detection. The problem is exacerbated by long-tailed, site-specific degradations and highly inconsistent illumination conditions, which limit the effectiveness of closed-set detectors such as ~\cite{fasterrcnn, sam2}. On the other hand, open vocabulary based methods perform poorly due to out of domain inputs.
\Cref{fig:qualitative} shows the output of three state-of-the-art open-vocabulary baselines (Lang-SAM~\cite{githubGitHubPaulguerrerolangsam}, Grounding DINO~\cite{grounding}, and Grounding SAM~\cite{groundingsam}) for three commonly used terms in inspection reports: rust, ice, and scaling. Grounding DINO yields coarse boxes; Grounding SAM converts those boxes to masks but spreads labels broadly. Further, fine-tuning existing models for this task is infeasible due to the lack of domain-specific data. The limited number of available human inspection reports are often qualitative and expressed in unstructured free text (see~\cref{fig:nycc-report}), lacking the consistency needed to support fine-tuning. 
Beyond visual detection, inspection methods that rely on 3D or photogrammetric reconstruction are also hindered by the same environmental factors. Poor and inconsistent illumination combined with a feature-sparse, confined geometry renders such reconstructions particularly unreliable. In these conditions, reconstruction algorithms often fail to capture micro-scale structural details, leading to smoothing over real faults like fine cracks. Prevailing challenges in photogrammetric and multi-view reconstruction include low-texture regions and uneven lighting, which systematically degrade geometric fidelity~\cite{rodriguez2025recent}. Moreover, quantitative studies confirm that insufficient surface texture introduces reconstruction noise and reduces accuracy, especially in fine-detail retention~\cite{nielsen2022quantifying}. In short, while robots can access these environments, perception and defect detection under challenging visual conditions are one of the major challenges for autonomous culvert inspection.

Large Vision–Language Models (VLMs) offer a promising alternative in this context. They have shown remarkable adaptability to vague or underspecified prompts, often outperforming hand-crafted baselines in zero- or few-shot settings~\cite{zhou2022learning}. Prompt engineering has been demonstrated to effectively steer VLMs across a wide range of tasks without fine-tuning, enabling robust generalization from loosely phrased instructions~\cite{Gu2023SurveyPromptEngineering}. Moreover, combining textual and visual prompting can compensate for the ambiguity or sparsity in any single modality, further enhancing performance in under-defined scenarios such as out-of-distribution segmentation~\cite{avogaro2025show}. When inspecting a culvert, a human does not carefully analyze every surface in detail from the start; instead, they take a quick glance to judge whether something looks unusual, and only then move closer to examine those regions more carefully. VLMs naturally mirror this behavior: they can be prompted with broad, loosely phrased queries to quickly highlight potentially relevant regions, and then refined with more specific prompts or visual cues to focus attention where it is most needed. 

Building on this analogy, we present \textbf{\qq{}} (Visual Inspection System with Intelligent Observation and Navigation), a language-in-the-loop inspection stack that runs onboard the robot. \qq{} couples vision–language reasoning with constrained viewpoint planning to enable autonomous inspection in culverts. 
Given a general prompt from an inspector and an image from a forward-facing camera, the VLM proposes candidate defect regions with bounding boxes, confidences, and natural-language rationales. 
These proposals drive a planner that accounts for culvert geometry and the robot’s limited mobility, commanding short forward motions and pan–tilt adjustments of a second camera to capture close-range, high-resolution imagery suitable for downstream measurement and audit.
Our contributions in the work are as follows:
\begin{itemize}
    \item We propose an end-to-end framework for fully autonomous culvert inspection 
    \item We demonstrate a methodology to leverage web-scale vision-language models zero-shot for an abstract inspection task
    \item We deploy our system onboard a legged robot with a pan-tilt gimbal, and validate its effectiveness through field experiments in an actual culvert.
\end{itemize}

\section{Related Work}
\label{sec:related}

\subsection{Vision-Language models for  planning}

The rise of foundation models has transformed high-level task planning in robotics. LLMs such as PaLM, GPT-3/4, and LLaMA can interpret natural language and decompose it into action sequences \cite{ahn2022can, huang2022inner}, while VLMs like CLIP and PaLM-E ground these plans in perception, linking what a robot “sees” to what it should “do” \cite{driess2023palm, brohan2023rt2}. Together, they enable high-level reasoning and common-sense inference anchored in physical affordances, pointing toward more generalist robotic agents.

A growing body of work applies these models to autonomous inspection, where robots must navigate complex environments, detect anomalies, and communicate findings. At the planning level, LLMs have been integrated into UAV inspection pipelines to interpret natural-language inspection goals and generate executable flight plans \cite{liu2024llm}. Closed-loop designs further allow LLMs to critique and refine drone trajectories, improving safety and reliability in sensitive inspection contexts \cite{wang2025large}. Complementarily, VLMs have been leveraged for text-guided inspection planning, identifying inspection targets from descriptions and optimizing trajectories in 3D environments \cite{sun2025text}. For visual understanding, CLIP-based systems enable zero-shot anomaly detection in industrial products \cite{qu2024vcp}. More advanced multimodal frameworks exploit GPT-4V’s reasoning to capture logical anomalies, such as missing or misconfigured components in assemblies \cite{zhang2025towards}. Finally, multimodal LLMs have been proposed for automated report generation, translating data collected by drones or ground robots into detailed inspection documents \cite{pu2024autorepo}. Collectively, these works highlight that foundation models provide both the cognitive layer (LLMs for sequencing, explanation, and interaction) and the perceptual grounding (VLMs for anomaly detection and semantic understanding) required for autonomous inspection, marking a shift from rule-based or task-specific pipelines to more flexible, general-purpose inspection systems.

\subsection{View Planning for Autonomous Inspection}
Autonomous inspection is often approached through coverage path planning (CPP), with 
more recent methods integrate sensing and uncertainty constraints, 
incorporating measurement uncertainty, optimizing structural coverage, 
and introducing roadmap-based inspection planning~\cite{Liu2022CPPUncertainty,Zhao2024StructuralInspection,Fu2021IRIS}.

In parallel, viewpoint optimization has been studied extensively in the context of active vision. 
Recent work has extended these ideas to inspection with pan–tilt–zoom (PTZ) cameras: \cite{Papaioannou2023IntegratedCPP} jointly optimize platform trajectory and gimbal orientation under visibility constraints, \cite{Wu2025PTZPlanning} ensure coverage while maintaining spatial resolution using PTZ-equipped robots, \cite{Ma2019FOVConstraints} derive explicit FOV constraints for pan–tilt cameras, \cite{arslan2018voronoi} develop gradient-based coverage controllers for PTZ networks. Similarly, \cite{wang2021visibility} formulate a differentiable visibility cost for joint position–orientation trajectory optimization. 

Another major thread is information-theoretic Next-Best-View (NBV) planning.
Modern variants employ sampling and learning to maximize information gain in inspection tasks \cite{gao2025take,naazare2022online,LHVP2024}. While effective for exploration, these methods are less suited to confined environments, where motion and visibility are highly constrained. 
In contrast, our work focuses on constrained coverage viewpoint planning in culverts, where motion is limited to axial translation with bounded pan–tilt actuation.

\section{\qq{}}
\label{sec:method}

\begin{figure}
    \centering
    \includegraphics[width=0.9\linewidth,trim={1.2cm 1cm 1.3cm 1.2cm},clip]{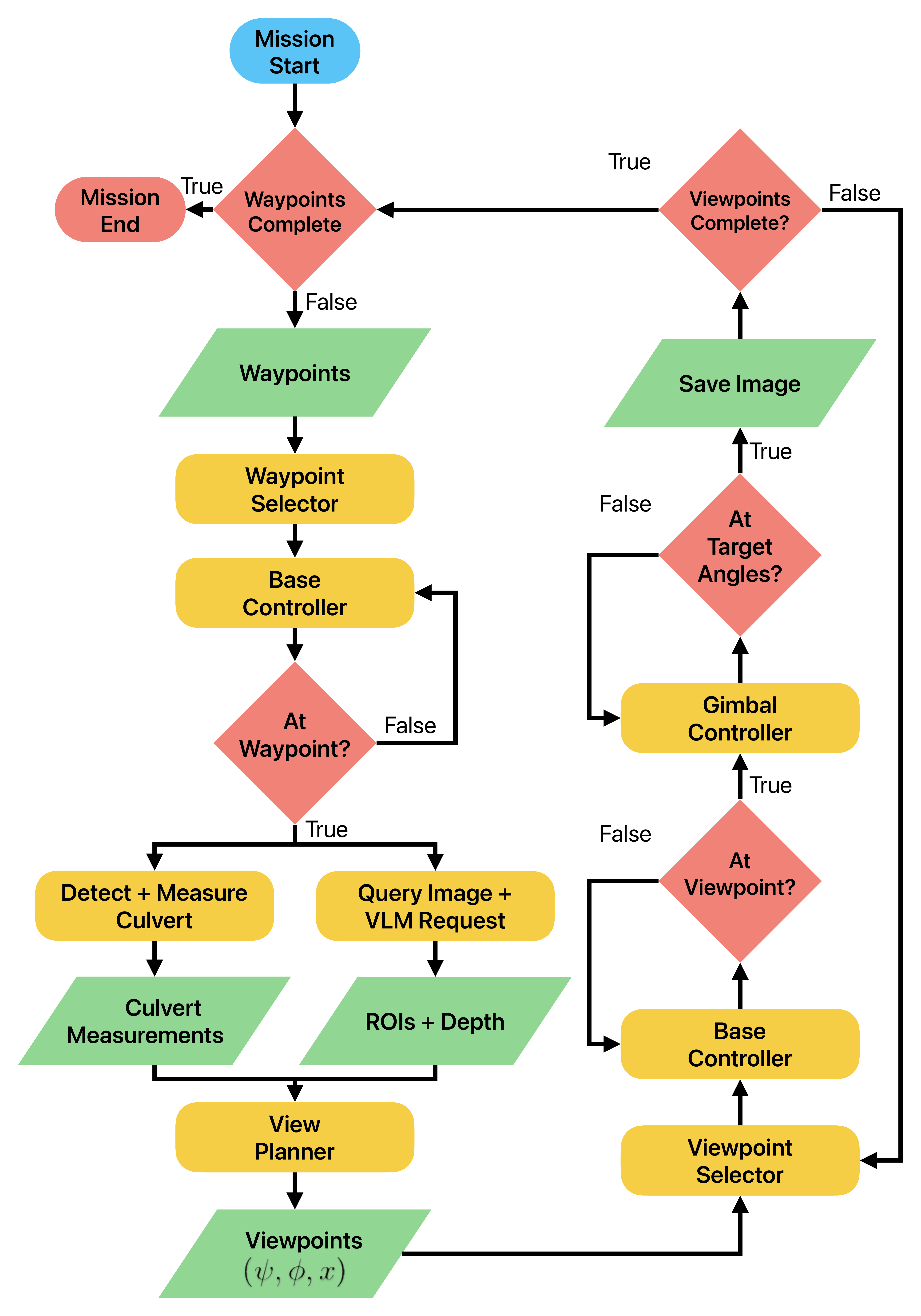}
    \caption{Overview of the \qq{} inspection pipeline. At each global waypoint, a query image is analyzed to extract ROIs and solve next-best viewpoints. Callout: for each ROI, the planner executes a local waypoint, commands the pan–tilt gimbal angles, and captures the inspection image. Then, moves on to the next waypoint}
    \label{fig:vision-method}
    \vspace{-0.5cm}
\end{figure}

This section describes \qq{}'s approach to inspection as shown in \cref{fig:vision-method}. We separate global navigation from navigation inside the culvert and assume that the robot can navigate globally to the entrance of the culvert. \qq{} is initiated when the robot is at the culvert opening. The robot is scheduled to visit waypoints along the culvert spaced every \textit{N} meters and perform two tasks - (i) detect and measure culvert geometry, and (ii) capture a query image that is sent to the VLM to obtain region of interest (ROI) proposals. This navigation and further reasoning happens in the Culvert Coordinate Frame (CCF) (described in \cref{sec:prelims}). The VLM-based ROI proposal methodology is described in \cref{sec:roi}. Once the ROIs for a segment of \textit{N} meters are received, a viewpoint optimization module computes the ideal viewpoints to capture images at every ROI proposed at a high resolution. For this, the module solves for viewpoints that provide requisite coverage of each ROI as well as the ideal viewpoint that can capture each ROI. These viewpoints are sorted by ascending order by x and executed by translating the robot in CCF to the correct position and rotating the gimbal appropriately each time. These two modules execute in tandem until all the waypoints in the culvert are visited. All collected images are then processed offline as described in \cref{sec:reporting}. 




\subsection{Region-of-Interest Proposal}
\label{sec:roi}
The region-of-interest (ROI) module queries a VLM to propose candidate ROIs. We use GPT-5 (via the OpenAI API) as the VLM and provide concise system and user prompts \footnote{We encourage the reader to visit this \href{https://sites.google.com/view/culvertinspectionwithvlms/home}{{\color{blue}page}} to review the prompts supplied to the VLM}. To avoid anchoring on a fixed defect taxonomy (e.g., corrosion, cracking), the prompts are intentionally general. In practice, this bare-bones prompting encouraged broader hypothesis generation and produced proposals that aligned more closely with expert assessments.

As shown in \cref{fig:vision-method}, ROI proposals from the VLM are fused with per-region depth derived from stereo. The VLM outputs normalized boxes $[x_{min},y_{min},x_{max},y_{max}]\in[0,1]$; we rescale them to pixel coordinates, sample the rectified depth map within each box, and compute summary depth statistics. This produces enriched ROIs that bundle the textual reason, a confidence score, the pixel-space bounding box, depth summaries (e.g., median range, spread/min–max), and 3D coordinates of the bounding boxes (obtained using the pinhole projection model together with the registered depth map). These attributes give the view planner information and context it needs to rank next-best viewpoints.
\subsection{Notations and Conventions}
\label{sec:prelims}
To reason consistently about robot motion and camera viewpoints inside the culvert,
we first establish a common spatial reference. Since culverts can be 
approximated as horizontal cylinders running beneath the canal, we define and 
relate three coordinate frames as follows.  

\textbf{Culvert Coordinate Frame (CCF):} Right-handed frame such that
$\mathbf{+x_\mathcal{S}}$ points along the culvert, $\mathbf{+z_\mathcal{S}}$ 
points upward (opposite gravity), and $\mathbf{+y_\mathcal{S}}$ completes the 
right-handed set. The robot (and thus the cameras) can translate only along $
\mathbf{x_\mathcal{S}}$, while its gimbal-mounted camera can yaw and pitch.

\textbf{Gimbal frame (}\(\mathcal{C}_G\)\textbf{):} Right-handed frame at the 
gimbal’s rotation center; $\mathbf{+z_\mathcal{G}}$ aligns with 
$\mathbf{+z_\mathcal{S}}$ (yaw axis), $\mathbf{+y_\mathcal{G}}$ is the pitch 
axis, and $\mathbf{+x_\mathcal{G}}$ points forward when $\psi=\phi=0$.  

\begin{figure}
    \centering
    \includegraphics[width=0.8\linewidth,trim={1cm 2cm 2cm 1cm},clip]{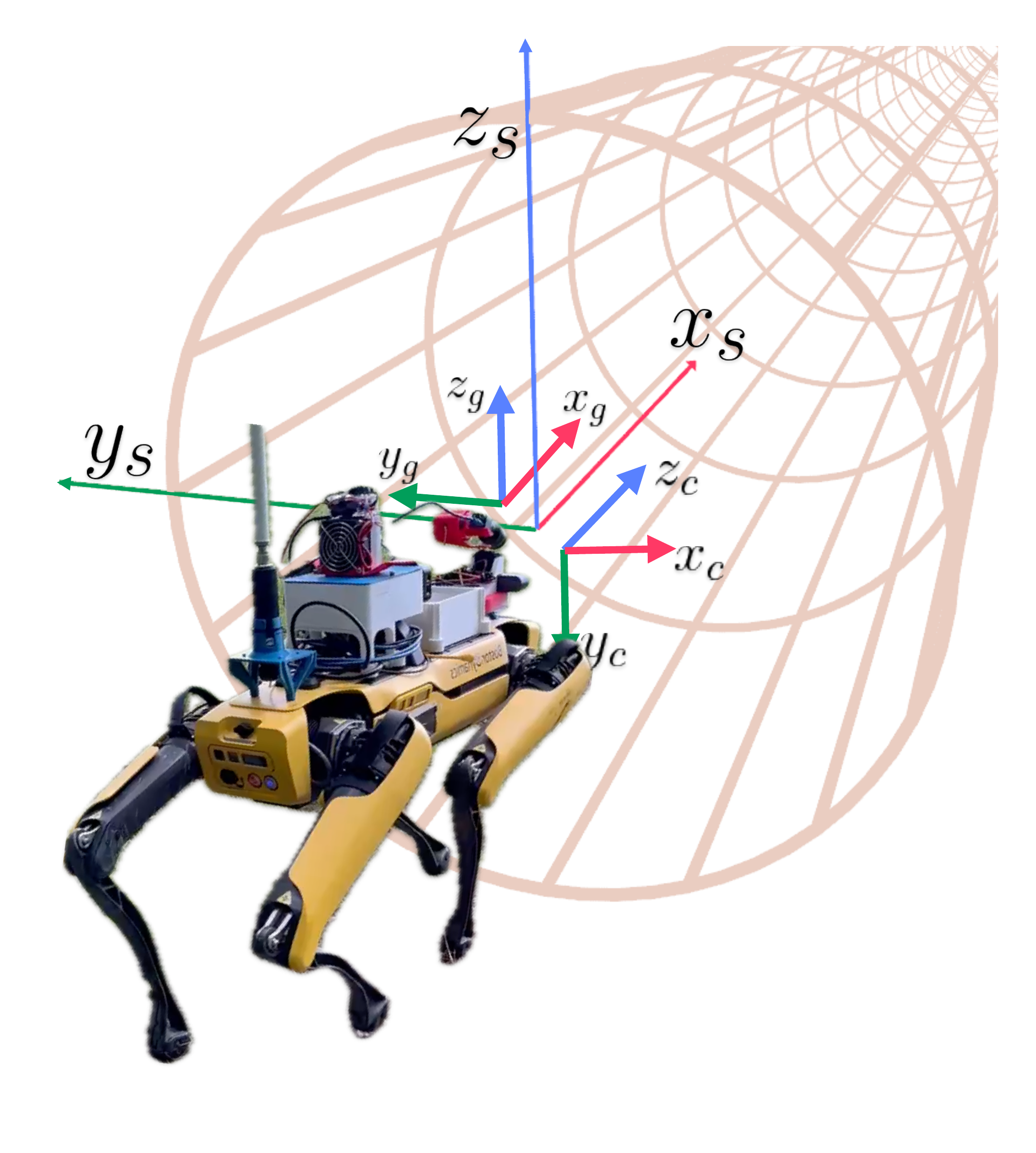}
    \caption{Coordinate frames used in our system: a culvert-fixed world frame$\{x_\mathcal{S},y_\mathcal{S},z_\mathcal{S}\}$ and the gimbal $\{x_\mathcal{G},y_\mathcal{G},z_\mathcal{G}\}$ and camera $\{x_\mathcal{C},y_\mathcal{C},z_\mathcal{C}\}$ frames mounted on the legged robot; axes colored x=red, y=green, z=blue. Gimbal axes are aligned with the world only in neutral position}
    \vspace{-0.8cm}
    \label{fig:ccs}
\end{figure}

\textbf{Camera frame (}\(\mathcal{C}\)\textbf{):} Right-handed frame fixed to 
each camera. By extrinsic calibration, $\mathbf{+z_\mathcal{C}}$ aligns with 
$\mathbf{+x_\mathcal{S}}$, $\mathbf{+x_\mathcal{C}}$ with $\mathbf{-y_\mathcal{S}}$, and $\mathbf{+y_\mathcal{C}}$ with $\mathbf{-z_\mathcal{S}}$. 
In our setup, two cameras are used: camera-1 , a fixed forward-looking camera for querying the VLM, and camera-2, a gimbal-mounted camera for capturing high-resolution images of the ROIs proposed by the VLM. The extrinsic calibration of both cameras is expressed with respect to the culvert coordinate frame, as illustrated in \cref{fig:ccs}. 

We also define the following transforms between these frames:
\begin{itemize}
    \item $\mathbf{T_{\mathcal{SC}_1}}$: Homogeneous transform from \textit{camera-1 to CCF}.
    \item $\mathbf{T_{\mathcal{SG}}^{(\bar{x})}}$: \textit{ Gimbal-to-CCF} homogeneous transform with $\bar{x}$ translation along the culvert axis $x_{\mathcal{S}}$. 



    \item The gimbal pans camera-2 about its local $+z_\mathcal{G}$ axis by angle $\psi$, with rotation matrix $\mathbf{R_z(\psi)}$ and tilts the camera about its local $+y_\mathcal{G}$ axis by angle $\phi$, with rotation matrix $\mathbf{R_y(\phi)}$

    \item $\mathbf{T_{\mathcal{GC}_2}}$: homogeneous transform from \textit{camera-2 to gimbal}.
    \item $\mathbf{T}_{\mathcal{SC}_2(\psi,\phi, \bar{x})}$: homogeneous transform from camera-2 to world by chaining the previous transforms as 
    \begin{align}
        \mathbf{T_{\mathcal{SC}_2(\psi,\phi, \bar{x})}} = T_{\mathcal{SG}}^{(\bar{x})} R_z(\psi)R_y(\phi)T_{\mathcal{GC}_2}
    \end{align}
    with, tilt (pitch) applied after panning (yaw).
\end{itemize}


\subsection{Viewpoint Optimization}
\label{sec:vp_opt}
The ROIs are obtained from a forward-looking camera, and might not accurately identify and highlight the fault as needed for reporting (See human report in \cref{fig:nycc-report}). 
A good viewpoint must ensure that the ROI lies fully within the camera’s field of view, is centered and scaled appropriately in the image, and is observed from a near-frontal orientation to preserve geometric fidelity. Such viewpoint optimization is performed on a second, higher resolution camera onboard our robot mounted on a pan-tilt gimbal for better maneuverability. However, the robot operates inside a narrow culvert where mobility is constrained to translation along the CCF x-axis, and the gimbal itself is bounded by mechanical pan-tilt limits and potential self-occlusion. As a result, not every desired viewpoint is physically reachable. Therefore, the objective of viewpoint optimization is to determine the camera-2 pose $C(\psi^*,\phi^*,\bar{x}^*)$ (in the CCF) that yields the best possible observation of the ROI subject to these constraints. Our approach, tailored specifically for constrained viewpoint planning in culverts, is detailed below.


We adopt a \emph{two-stage strategy} because the objective landscape is highly non-convex with many local minima. In \emph{stage one, a coarse brute-force grid search} over the bounded yaw, pitch, and translation is used to obtain a feasible seed configuration. This discrete search avoids poor local minima and ensures full coverage of the configuration space. The discretization for this is empirically selected to ensure tractability in practice, such that solving for a single viewpoint takes on the order of 5 to 8 seconds. This seed is then \emph{refined using a gradient-based optimization in stage two}, which converges rapidly to a optimal solution in 15-20ms.

First, the 3D bounding box coordinates obtained in the camera-1 frame (\cref{sec:roi}) are transformed into the CCF using the extrinsic calibration $T_{\mathcal{SC}_1}$: 
\begin{align}
    p_\mathcal{S} = T_{\mathcal{SC}_1}\,p_{\mathcal{C}_1} \notag
\end{align}
\noindent Then the ROI center and its local surface normal are approximated from the cylindrical geometry of the culvert. To robustly estimate the culvert diameter, a circle is fitted to the local point cloud using RANSAC, and the ROI points are radially clipped to lie on the estimated cylinder.

Next, these points are transformed from the CCF into the camera-2 frame via
\begin{align}
    p_{\mathcal{C}_2} = T_{\mathcal{C}_2S}\,p_\mathcal{S}, 
    \qquad T_{\mathcal{C}_2S} = T_{\mathcal{SC}_2}^{-1}, 
\end{align}
and projected into the image plane using camera-2’s intrinsics $K$:  
\begin{equation}
    \begin{bmatrix}
        u_i \\ v_i \\ 1
    \end{bmatrix}
    =
    K
    \begin{bmatrix}
        X_i/Z_i \\ Y_i/Z_i \\ 1
    \end{bmatrix}, 
    \qquad
    K =
    \begin{bmatrix}
        f_x & 0 & c_x \\
        0 & f_y & c_y \\
        0 & 0 & 1
    \end{bmatrix}
\end{equation}
This yields the 2D image coordinates $(u_i,v_i)$ of the ROI vertex $i$ in camera-2’s view. The goal is to determine the camera-2 pose that projects the ROI such that each vertex $(u_i, v_i)$ lies fully within the field of view (\cref{eq:coverage}), while the ROI as a whole remains centered near the principal point (\cref{eq:centering}), scaled to an appropriate size (\cref{eq:size}), observed from a near-frontal orientation to preserve geometric fidelity (\cref{eq:obli}) and in focus (\cref{eq:focus}).
\\ \\
\noindent \textbf{Stage 1: Coarse grid search}\\ The search evaluates the 
objective $\mathbf{J_1}$ over a discretized set of yaw, pitch, and translation 
values, and returns the configuration $(\psi_0, \phi_0, x_0)$ that attains the 
minimum cost within the specified bounds.
\begin{equation}
    (\psi_0, \phi_0, \bar{x_0}) = 
    \underset{\psi,\phi,\bar{x}}{\arg\min}\; J_1(\psi,\phi,\bar{x}),
\end{equation}
Where, 
\begin{equation}
    \begin{aligned}
        \mathbf{J_1}(\psi,\phi,\bar{x}) &=
        w_{\mathrm{cov}}\,L_{\mathrm{cov}}
        + w_{\mathrm{ctr}}\,L_{\mathrm{ctr}}
        + w_{\mathrm{size}}\,L_{\mathrm{size}} \\
        &\quad+ w_{\mathrm{obl}}\,L_{\mathrm{obl}}
        + w_{\mathrm{range}}\,L_{\mathrm{range}} \, 
\end{aligned}
\label{eq:J1}
\end{equation}
\noindent here, \\
\noindent The coverage cost penalizes ROI points that leave the image, ensuring full visibility
\begin{equation}
    \begin{aligned}
        L_{\mathrm{cov}} = \sum_{i=1}^M \Big[
        \operatorname{softplus}(m - u_i) +
        \operatorname{softplus}(m - v_i) + \\
        \operatorname{softplus}(u_i - (W-m)) +
        \operatorname{softplus}(v_i - (H-m))
        \Big],
    \end{aligned}
    \label{eq:coverage}
\end{equation}
with margin $ m = \alpha \cdot \min(W,H)$, $\alpha \in$ [0,1] is a scalar that specifies the required distance of the ROI points from the image boundaries.

\noindent The centering cost encourages the projected ROI centroid to remain close to the image principal point, reducing off-axis views.
\begin{equation}
L_{\mathrm{ctr}} = (\bar{u} - c_x)^2 + (\bar{v} - c_y)^2,
\quad
(\bar{u},\bar{v}) = \tfrac{1}{M}\sum_{i=1}^M (u_i,v_i)
\label{eq:centering}
\end{equation}

\noindent The size cost regulates the projected area of the ROI, encouraging it to occupy a target fraction of the image.
\begin{equation}
L_{\mathrm{size}} = \Big[(u_{\max}-u_{\min})(v_{\max}-v_{\min})
- f_{\mathrm{target}}WH\Big]^2 
\label{eq:size}
\end{equation}

\noindent The obliquity cost penalizes grazing views by encouraging alignment between the camera optical axis and the ROI surface normal.
\begin{equation}
L_{\mathrm{obl}} = \big(1 - |z_{\mathrm{cam}}\cdot n_c|\big)^2,
\quad z_{\mathrm{cam}} = R(\psi,\phi)[0,0,1]^\top 
\label{eq:obli}
\end{equation}

\noindent The range cost enforces a desired standoff distance ($d^*$) between the camera center ($C(\psi,\phi,\bar{x})_c$) and the ROI center ($ROI_c$) in the world frame. This is ensures focus and avoids blurry images.
\begin{equation}
L_{\mathrm{range}} =
\Big(||ROI_c - C(\psi,\phi,\bar{x})_c|| - d^*\Big)^2 
\label{eq:focus}
\end{equation}

\noindent \textbf{Stage 2: Gradient based refinement} \\ The configuration obtained from the grid search is refined by minimizing the objective $\mathbf{J_2}$  using the quasi-Newton solver L-BFGS-B:  
\begin{equation}
    \begin{aligned}
        \mathbf{J_2}(\psi,\phi,\bar{x}) = J_1 
        + w_{\mathrm{move}}\,L_{\mathrm{move}}
        + w_{\mathrm{trans}}\,L_{\mathrm{trans}}.
    \end{aligned}
\end{equation}
Where, $J_1$ is computed as described previously in~\cref{eq:J1}.
\noindent The motion cost discourages large deviations in yaw and pitch from the seed values:
\begin{equation}
    L_{\mathrm{move}} = [(\psi-\psi_0)/ \Delta_\psi]^2 +
    [(\phi-\phi_0)/\Delta_\phi)]^2,
\end{equation}
where $\Delta_\psi$ and $\Delta_\phi$ are tolerances for deviations in yaw and pitch, respectively.  

\noindent The translation cost discourages large displacements of the camera from $\bar{x}_0$:
\begin{equation}
    L_{\mathrm{trans}} = [(\bar{x}-\bar{x}_0)/\Delta_x]^2,
\end{equation}
where $\Delta_x$ is the tolerance for deviation.  

We do this 2-stage optimization for each ROI, resulting in a set of optimal camera-2 poses that 
enable consistent, high-quality re-imaging under the culvert’s geometric and 
mechanical constraints.


\subsection{Reporting and Assessment}
\label{sec:reporting}
As a final step, close-up images captured at the viewpoints are re-evaluated by the VLM using an assessment prompt. For each ROI, we supply the original hypothesis along with its follow-up crop(s). The VLM then evaluates whether the hypothesis is supported. It outputs a JSON record containing the original rationale, a result label $\in$ \{{\texttt{Confirmed, Partially Confirmed, Not Confirmed}\}, and, when applicable, a one-line defect description. These descriptions are used for qualitative evaluations as described in \cref{sec:eval}. 

\section{Experimental Evaluation}
\label{sec:eval}


\subsection{Experiment Setup}

The platform was a Boston Dynamics Spot outfitted with a compact custom pan–tilt gimbal providing 120° pan and 45° tilt. A ZED 2i (camera-1) provided proposal imagery and stereo depth. A FLIR Blackfly (camera-2), co-located with a dimmable inspection light, captured high-resolution follow-ups. A 50 W flood LED supplied general illumination.
We conducted multiple autonomous runs in a culvert beneath the Erie Canal. For each run, a small set of global waypoints was given to \qq{}, which navigated Spot to each waypoint. At each stop, our pipeline (\cref{fig:vision-method}) proposed ROIs, fused depth, and estimated the local culvert diameter. The planner then used these estimates to compute the best viewpoints for targeted re-imaging.

The stack is implemented in ROS 2 with Scipy for optimization and the OpenAI API for VLM queries, running online on an onboard NVIDIA Jetson. To support cloud calls from within the culvert, we deployed a point-to-point (PtP) Ubiquiti radio link that provided long-range connectivity for reliable OpenAI API requests. We designed the gimbal to support pan (yaw) and tilt (pitch) while carrying the camera and co-located light within Spot’s footprint. Both light sources are fully ROS-controllable and are only actuated during image capture to conserve power. For state estimation, we use Spot’s odometry and republish it into the CCF. All components are packaged as ROS 2 actions, orchestrated by a finite-state behavior controller.

We conducted 8 autonomous runs in Culvert 110 (Gasport, NY)—a 66 m-long, 1.2 m-diameter conduit running roughly north–south beneath the Erie Canal. The environment is extremely low-light, uneven, and typically contains 6–10 in of standing water. Our robot, running \qq{}, operated end-to-end: capture wide images every 5 meters → query the VLM → obtain ROI proposals → plan viewpoints → acquire close-up re-images. The initial proposal frame and the follow-up close-ups were then returned to the VLM to verify or revise the original hypotheses.


\begin{table}[]
\begin{tabular}{@{}c|l|lll|lll@{}}
\toprule
\multicolumn{1}{l|}{} &  & \multicolumn{3}{l|}{\begin{tabular}[c]{@{}l@{}}Non-Author \\ Evaluator (NAE)\end{tabular}} & \multicolumn{3}{l}{\begin{tabular}[c]{@{}l@{}}Subject-Matter \\ Expert (SME)\end{tabular}} \\ \midrule
\multicolumn{1}{l|}{} & Scene & RAR & CC & DDS & RAR & CC & DDS \\ \midrule
\multirow{5}{*}{\begin{tabular}[c]{@{}c@{}}ROI \\ +\\ Description\end{tabular}} & 1 & 0.958 & 0.5 & 2.5/3 & 0.667 & 1 & 2.5/3 \\
 & 2 & 0.813 & 0.625 & 2.375/3 & 1 & 1 & 2.5/3 \\
 & 3 & 0.708 & 0.625 & 2.625/3 & 0.833 & 1 & 2.5/3 \\
 & 4 & 0.75 & 0.625 & 2.5/3 & 0.5 & 1 & 2/3 \\
 & 5 & 0.833 & 0.72 & 2.875/3 & 0.5 & 1 & 2.5/3 \\ \midrule
\multirow{3}{*}{ROI Only} & 6 & 0.687 & 0.625 & \multirow{3}{*}{N/A} & 0.75 & 0.5 & \multirow{3}{*}{N/A} \\
 & 7 & 0.5 & 0.625 &  & 0.667 & 1 &  \\
 & 8 & 0.708 & 0.5 &  & 0 & 1 &  \\ \bottomrule
\end{tabular}
\caption{Shows Reasoning-agreement rate (RAR), coverage confidence (CC) and defect-description score (DDS) on a scale of 1-3. NAEs report a mean 74.4\% RAR, 60.4\% CC and 2.575 DDS while SMEs report a mean 61.4\% RAR, 93.75\% CC and 2.4/3 DDS.}
\vspace{-0.7cm}
\label{tab:results}
\end{table}

\subsection{Defect-detection Results}
\qq{}’s performance was assessed by non-author evaluators (NAEs) and subject-matter experts (SMEs) from the New York Canal Corporation (NYCC). The NAEs were made to read an official NYCC inspection report prior to answering the survey in order to provide them with context. 
We surveyed 8 NAEs and 2 SMEs and report three metrics: \textbf{reasoning-agreement rate (RAR)}—the fraction of proposed ROIs whose reasoning evaluators agreed with; \textbf{coverage confidence (CC)}—the fraction of scenes for which evaluators believed the VLM missed no significant defects; and \textbf{defect-description score (DDS)}—a quality rating of the VLM’s final one-line degradation descriptions.


Using these metrics, we assessed ROI proposals on 8 scenes and final defect-descriptions on 5 scenes. \Cref{tab:results} summarizes the results: the initial ROI proposals from \qq{} achieved \textbf{67.96\% }RAR agreement with human judgments, and the final, post–re-imaging assessments reached \textbf{2.48} DDS (scale of 1-3, where 3 is the best), highlighting \qq{}’s ability to produce coherent narratives of defects.

The numbers in \cref{tab:results} paint a nuanced picture. On average NAEs were more likely to accept the VLM’s reason for proposing the boxes (mean RAR = \textbf{74.4\%}) than SMEs (mean RAR = \textbf{61.4\%}), but SMEs expressed far greater confidence that the system’s outputs covered all meaningful defects (mean CC \textbf{93.8\%} for SMEs vs \textbf{60.4\%} for NAEs).  Quality of the one-line descriptions was consistently reasonable: NAEs gave a mean DDS of \textbf{2.575/3} and SMEs \textbf{2.4/3}, and the post–re-imaging DDS (\textbf{$\approx$2.48}) reported earlier is consistent with these values. Taken together, these results suggest the VLMs generates coherent, useful defect narratives that often compensate for imperfect boxes (hence high SME coverage confidence despite lower RAR). Importantly, NAEs adopt a more conservative stance than SMEs: they are quick to agree with the box reasoning because it aligns with common-sense intuition, yet remain skeptical that all defects have been captured. SMEs, by contrast, apply domain-specific judgment to filter spurious boxes more effectively and are more confident that the few critical defects of interest have already been identified by the VLM. This contrast suggests a valuable role for non-domain experts in inspection: NAEs can act as a “step 0,” conducting preliminary reviews and flagging candidate defects before SMEs evaluate the results, thereby easing the workload on domain specialists. This is made possible by the VLM’s natural-language descriptions, which present findings in a form that non-domain experts can readily interpret.

Qualitatively, the defect highlighted in ROI 1 of~\cref{fig:results} was identified by all SMEs as significant. It corresponds to the most frequently reported issue in inspection records: the joint between between older culvert and the extension (\cref{fig:nycc-report}). To evaluate whether the VLM consistently detects this defect, we queried it with five different images of the same junction. Across all five cases, the VLM identified the same region of interest (ROI), demonstrating consistent performance in recognizing this critical defect. This outcome aligns with the SMEs’ CC metric being consistently 1 for this case. Conversely, the ROIs whose hypotheses were not confirmed during the assessment step were precisely those lacking SME agreement (thus the lower RAR), further indicating that automated assessment decisions track expert consensus.
\begin{figure}[htbp]
    \vspace{-0.5cm}
    \centering
    \includegraphics[width=1\linewidth]{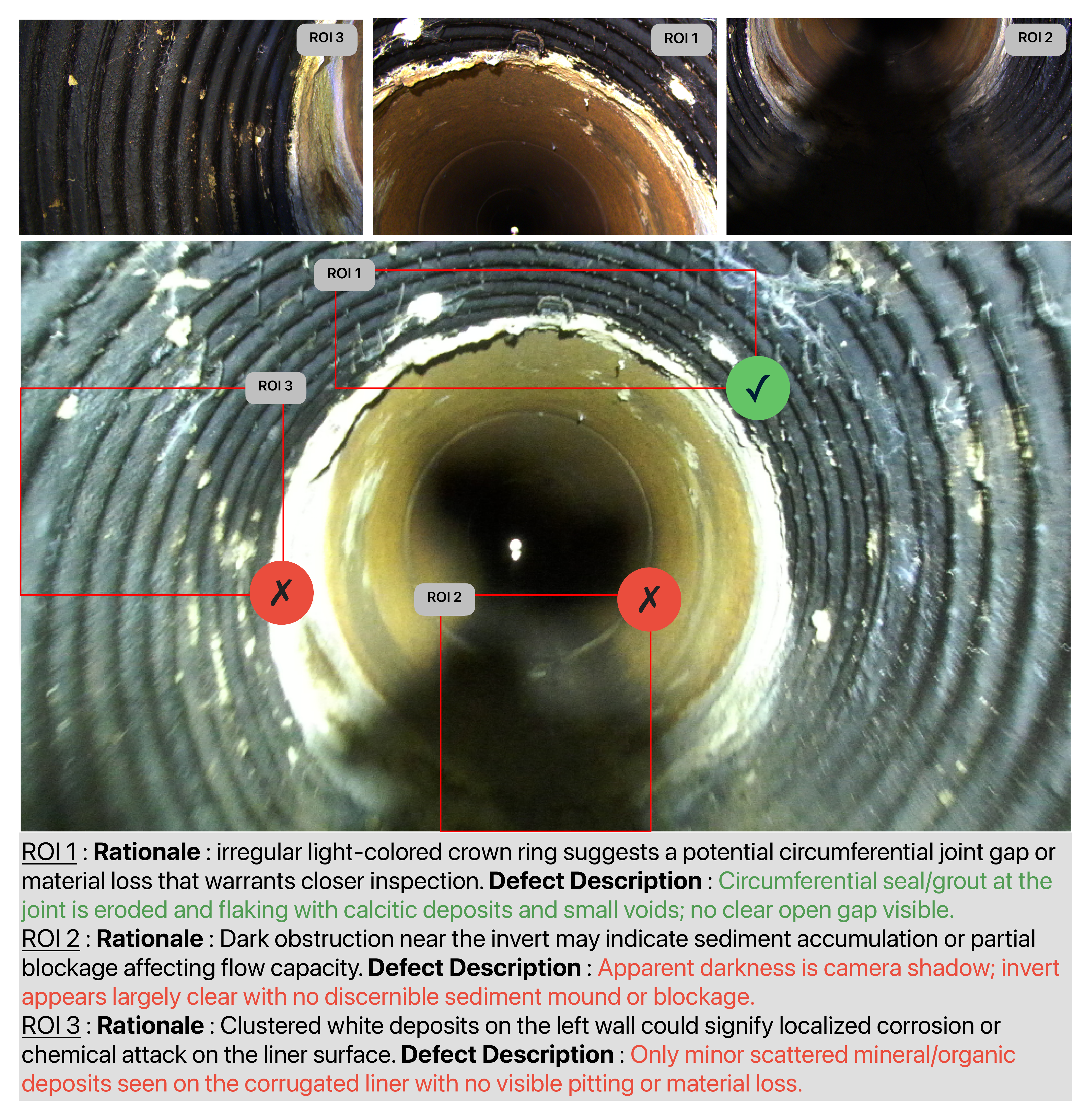}
    \caption{VLM ROI proposals on a query image frame (ROIs 1,2,3) with rationales and descriptions}
    \label{fig:results}
    \vspace{-0.85cm}
\end{figure}
\subsection{Runtime Summary}
Capturing a wide-frame image, uploading it to the VLM, and receiving ROI proposals took on average less than 60 s. Each global waypoint produced 3–4 ROIs, with batched viewpoint optimization  completing in under 20 s on the onboard Jetson. Executing a single ROI visit—including axial motion, pan–tilt adjustment, and a 5 s stabilization pause during which the co-located inspection light is activated—required roughly 15 s. In practice, a mission with three global waypoints spaced 5 m apart completed in about 7 min. Extrapolating to the full 66 m culvert, the system can complete an inspection in under 90 min including setup, compared to approximately 2 h for a human inspector—demonstrating that \qq{} offers greater efficiency.



\section{Discussion}
\label{sec:disc}




\noindent\textbf{Considerations of Language-in-the-loop Inspection}: Our VLM-based pipeline enables safer, more frequent culvert inspections that were previously limited by confined-space hazards and cost. Its open-vocabulary ROI discovery mirrors human intuition—scanning a scene with minimal guidance to ask “what looks off?”—and finding candidate degradations without relying on a fixed defect taxonomy, removing the heavy burden of collecting task-specific datasets and training rigid, closed-set detectors. Instead, inspectors can simply prompt the VLM, with every call operating as a fresh query that flexibly adapts to the scene. This makes autonomy far more accessible to domain experts, as the VLM effectively bridges human expertise and robotic perception. Instead of grappling with the technical complexity of dataset design, model training, or algorithm tuning, experts can express their knowledge in natural prompts that the VLM translates into actionable perception outputs. In doing so, the VLM bridges the gap between human intuition and robotic capability, lowering the barrier to deploying advanced autonomy in real inspection settings. Complementing this, data from NAEs indicates that robotics and autonomy experts can conduct preliminary inspections and vet results before they reach SMEs, further reducing the workload on domain specialists.
\\ 
\noindent\textbf{Limitations and future work}: Long-term deployment is constrained by two factors. First, the pipeline currently depends on networked VLM calls, which restricts operation in low-connectivity environments. Second, viewpoint selection is bounded by the pan–tilt gimbal’s mechanical limits and occasional self-occlusion from the robot body inside the culvert. Going forward, we plan to run compact, locally hosted VLMs fully on-board to remove the connectivity bottleneck and to auto-generate comprehensive inspection reports; we will also explore planner/gimbal co-design to expand reachable view angles and reduce occlusion.
Another avenue for future is to extend the system toward regular field deployment, enabling weekly or monthly inspections of the same culvert segments. Such scheduled operation would allow the robot to autonomously revisit previously identified defects, monitor their progression over time, and flag new anomalies as they appear. This would provide a reliable and higher-frequency monitoring pipeline, turning one-off inspections into a long-term asset management strategy.

\section{Conclusion}
\label{sec:conc}
We presented \qq{}, an autonomous culvert–inspection pipeline that couples a web-scale VLM with depth-aware, constraint-respecting view planning. From brief prompts, the VLM proposes open-vocabulary ROIs, which we enrich with stereo depth and use to select next-best viewpoints that avoid base yaw and minimize reverse motion. Deployed on a quadruped in an Erie Canal culvert, \qq{} closed the see→decide→move→re-image loop. Across eight ROI scenes and five description scenes, evaluators reported strong performance, and qualitative results showed clearer, more localized findings. Limitations include reliance on networked VLM calls, small sample size, and subjective scoring. Future work targets fully on-board models, larger evaluations with objective ground truth, and auto-generated inspection reports. Overall, \qq{} turns broad prompts into verified, expert-aligned findings for safer, higher-fidelity culvert inspection.

\bibliographystyle{ieeetr} 
\bibliography{misc} 

\end{document}